%% file: emnlp2022.tex
\pdfoutput=1
\documentclass[11pt]{article}

\usepackage[]{EMNLP2022}

\usepackage{times}
\usepackage{latexsym}

\usepackage[T1]{fontenc}
\usepackage[utf8]{inputenc}

\usepackage{microtype}

\usepackage[para]{threeparttable}
\usepackage{booktabs}
\usepackage{multirow}
\usepackage{graphicx}
\usepackage{amsmath}
\usepackage{amssymb}
\usepackage{hyperref}
\usepackage{mathtools}
\usepackage[ruled,vlined]{algorithm2e}

\input{commands}

\newcommand\blfootnote[1]{%
  \begingroup
  \renewcommand\thefootnote{}\footnote{#1}%
  \addtocounter{footnote}{-1}%
  \endgroup
}

\usepackage{inconsolata}

\hypersetup{
    breaklinks=true,
    colorlinks=true,
    pdfusetitle=true,
}

\title{Generating Information-Seeking Conversations from Unlabeled Documents}

\author{
    Gangwoo Kim$^{1 \clubsuit *}$ \quad Sungdong Kim$^{2,3*}$ \quad Kang Min Yoo$^{2,4}$ \quad Jaewoo Kang$^{1\dagger}$ \\
    Korea University$^{1}$ NAVER AI Lab$^{2}$ KAIST AI$^{3}$  NAVER CLOVA$^{4}$ \\
    \texttt{\{gangwoo\_kim, kangj\}@korea.ac.kr} \\
    \texttt{\{sungdong.kim, kangmin.yoo\}@navercorp.com}
}

\begin{document}
\maketitle

\blfootnote{\textsuperscript{$\clubsuit$} Work done while interning at NAVER AI Lab}
\blfootnote{\textsuperscript{$\ast$} Equal contribution \textsuperscript{$\dagger$} Corresponding author}
\input{tabs/00_abstract}

\input{tabs/01_introduction}

\input{tabs/02_background}

\input{tabs/03_method}

\input{tabs/04_experements}

\input{tabs/05_analysis}

\input{tabs/06_wiki}

\input{tabs/07_related_works}
\input{tabs/08_conclusion}

\input{tabs/99_limitations}

\input{tabs/acknowledgement}

% \section*{Acknowledgements}

\bibliography{anthology,custom}
\bibliographystyle{acl_natbib}

\appendix

\clearpage

\input{tabs/09_appendix}

\end{document}

%% file: commands.tex
\definecolor{purple1}{HTML}{7570b3}
\definecolor{green1}{HTML}{1b9e77}
\definecolor{orange}{HTML}{d95f02}
\definecolor{green2}{HTML}{4daf4a}
\definecolor{red}{HTML}{e41a1c}
\definecolor{blue}{HTML}{377eb8}
\definecolor{purple2}{HTML}{984ea3}

%% file: tabs/00_abstract.tex
\begin{abstract}

Synthesizing datasets for conversational question answering (CQA) from unlabeled documents remains challenging due to its interactive nature.
Moreover, while modeling \textit{information needs} is an essential key, only few studies have discussed it.
In this paper, we introduce a novel framework, \textsc{SimSeek}, (\textbf{Sim}ulating information-\textbf{Seek}ing conversation from unlabeled documents), and compare its two variants.
In our baseline \textsc{SimSeek-sym}, a questioner generates follow-up questions upon the predetermined answer by an answerer.
On the contrary, \textsc{SimSeek-asym} first generates the question and then finds its corresponding answer under the conversational context.
Our experiments show that they can synthesize effective training resources for CQA and conversational search tasks.
As a result, conversations from \textsc{SimSeek-asym} not only make more improvements in our experiments but also are favorably reviewed in a human evaluation.
We finally release a large-scale resource of synthetic conversations, \textsc{Wiki-SimSeek}, containing 2 million CQA pairs built upon Wikipedia documents.
With the dataset, our CQA model achieves the state-of-the-art performance on a recent CQA benchmark, QuAC \cite{choi2018quac}
\footnote{The code and dataset are available at \href{https://github.com/naver-ai/simseek}{github.com/naver-ai/simseek}.}.

\end{abstract}

%% file: tabs/01_introduction.tex
\section{Introduction}

\input{figs/example}

%**[Introduction to CQA]**
Conversational question answering (CQA) involves modeling the information-seeking process of human dialogue.
In the task, systems should understand questions according to the conversational context.
To build robust systems, large-scale CQA datasets \cite{choi2018quac, reddy2019coqa, saeidi2018interpretation, campos2020doqa} have recently been introduced. Still, they are limited in scale to generalize toward real-world applications, which motivates the development of automated methods for constructing CQA datasets.

However, generating CQA datasets is a challenging task, which requires interactions between questioner and answer.
Therefore, most of the literature has discussed subparts of the overall process.
One line of research in conversational question generation (CQG) aims to generate follow-up questions upon held-out conversations~\cite{pan2019reinforced, qi2020stay, gu2021chaincqg}.
Another line of research in CQA has greatly enhanced answer accuracy~\cite{qu2019attentive, kim2021learn, zhao2021ror}. 
Despite the recent advances, they assume all other ingredients (i.e., gold history by humans) are given.

Moreover, modeling \textit{information needs} can facilitate simulating realistic conversations.
As illustrated in Figure \ref{fig:example} (a),
% , the question are not coherent to the conversation.
questioners with excessive information often ask questions incoherent with the conversation.
On the other hand, the \textit{information needs} drive the questioners to seek new knowledge via conversation, failing to do so sometimes, as shown in Figure \ref{fig:example} (b).
However, only few CQA studies have focused on it~\cite{qi2020stay}.
% Nevertheless, most studies have been discussing the problem of generating questions based on answerers' knowledges, following conventional generation frameworks proposed for the single-turn QA task \cite{alberti2019synthetic, puri2020training, lewis2021paq}.

In this paper, we propose a novel framework, \textsc{SimSeek} (\textbf{Sim}ulating Information-\textbf{Seek}ing conversation from unlabeled documents) and compare its two variants. Both consist of two sub-modules, questioner and answerer, that converse with each other; but each variant assumes opposite scenarios, respectively.
% assuming each scenario, respectively. 
In (1) \textsc{SimSeek-sym}, an answerer first identifies answers from the document, and then a questioner asks context-dependent questions based on the predetermined answer.
% , which are automatically provided in advance by an extractive model.
% Although it succeeds in generating reasonable questions, 
On the contrary, (2) \textsc{SimSeek-asym} allows a questioner to ask questions without any prior knowledge about the answer.
Then, an answerer provides corresponding answers to the asked questions.
Either way, \textsc{SimSeek} sequentially generates QA pairs at every turn, moving the conversation forward.

% 단순 순서의 차이일지라도 결과는 사뭇 다르다
We generate synthetic conversations with our frameworks and evaluate them.
Despite the similarity of the two frameworks, \textsc{SimSeek-asym} performs better, which reveals the importance of modeling \textit{information needs}.
We first conduct experiments on a challenging CQA benchmark, QuAC~\cite{choi2018quac} in the semi-supervised setup.
Our experimental results demonstrate the effectiveness of the synthetic dataset from \textsc{SimSeek-asym} that outperforms other CQA generation baselines.
Besides, it also enhances the dense retriever on a conversational search benchmark, OR-QuAC \cite{qu2020open}.
% ** [Analysis] **
% To provide a deeper perspective into the information-seeking behavior, we thoroughly analyze and compare the resulting conversations from two frameworks. 
We perform a human evaluation to investigate how different conversations are generated by two variants, compared to the original human ones.
% We perform human evaluation to investigate how our two frameworks work differently and also compare them with human annotation.
As a result, conversations from \textsc{SimSeek-asym} are more favorably reviewed in overall adequacy than others, including humans.

We finally construct a large-scale resource of synthetic conversations, \textsc{Wiki-SimSeek}, which contains 2 million CQA pairs built upon 213k Wikipedia passages.
Further trained on the dataset, our CQA model achieves state-of-the-art performance on QuAC.
We hope it would shed light on building more robust CQA models and identifying the factors for realistic information-seeking conversation.

Our main contributions are summarized as:
% **[Contributions]**
\begin{itemize}
    \setlength\itemsep{0em}
  \item We propose a novel framework \textsc{SimSeek} that generates synthetic conversations from unlabeled documents and compare its two variants to provide a deeper understanding of the information-seeking conversation.
  \item To the best of our knowledge, we are the first to demonstrate the effectiveness of synthetic datasets in two downstream tasks, CQA and conversational search.
  \item We construct and release a large-scale resource of synthetic conversations, \textsc{Wiki-SimSeek}. By leveraging it, we achieve state-of-the-art performance on a challenging CQA benchmark, QuAC.
%   successfully adapt the SimSeek to semi-supervised CQA for the first time.
\end{itemize}

%% file: figs/example.tex
\begin{figure} [t!]
    \centering
    \includegraphics[width=\columnwidth]{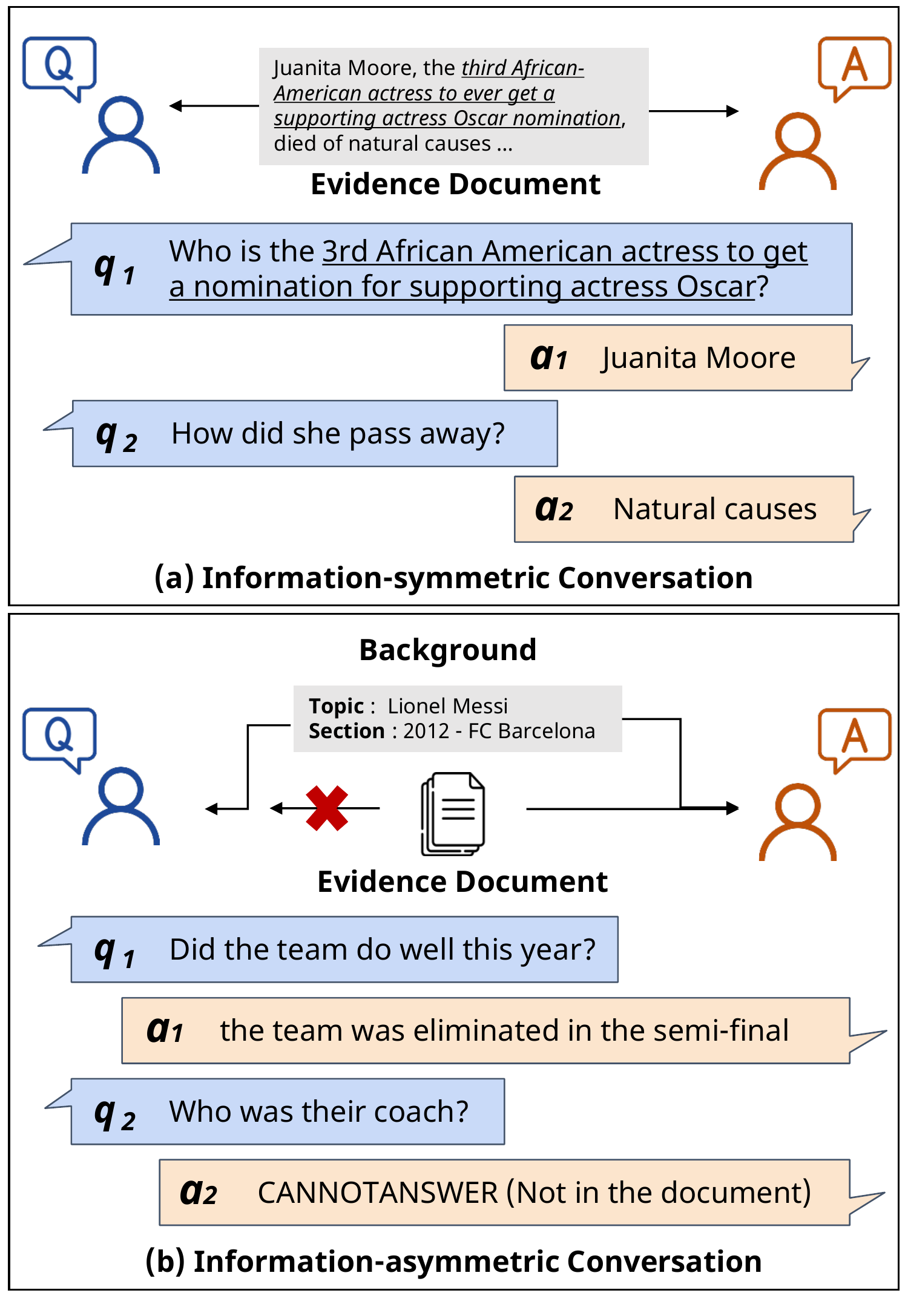}
    \caption{Examples of two conversation scenarios. In the former, questioners with access to the evidence document often ask less related things to the conversation. In the latter, questioners seek new information from the inaccessible document, leading to information-seeking behaviors, i.e., open-ended and unanswerable questions.}\vspace{-0.4cm}
    \label{fig:example}
\end{figure}

%% file: tabs/02_background.tex
\section{Background}
\input{figs/overview}

In the information-seeking conversation, two agents (i.e., questioner and answerer) converse about the specific topic. 
To provide accurate knowledge to the questioner, the answerer can utilize the document that consists of answer-containing passage $c$ and its background knowledge $\mathcal{B}$ (i.e., the title and abstract). 
Let $q_t$ be the current question and $a_t$ be its corresponding answer at turn $t$.
Formally, CQA systems are required to find correct answer $a_t$ to the question $q_t$ from the passage $c$ based on the conversational history $\mathcal{H}_{<t} = [(q_1,a_1),...,(q_{t-1},a_{t-1})]$, i.e. $p(a_t \mid q_t, c, \mathcal{H}_{<t})$.

Typically, most CQG research assumes that the questioner can access the answer-containing passage $c$. 
Hence, they formulate the task of generating the question $q_t$ based on the passage $c$ and answer $a_t$, i.e. $p(q_t \mid c, a_t, \mathcal{H}_{<t})$~\cite{gao2019interconnected, pan2019reinforced, gu2021chaincqg}.
The formulation can be considered a straightforward extension of the dominant paradigm in single-turn QA generation \cite{puri2020training, lewis2021paq}, where models generate questions given their answers.
Recently, \citet{qi2020stay} suggest a new viewpoint to promote a natural scenario of information-seeking conversation. 
In the setup, CQG modules are blinded to the answer-containing passage and rely on background information $\mathcal{B}$ when generating conversational question $q_t$, i.e. $p(q_t \mid \mathcal{B}, \mathcal{H}_{<t})$.

%% file: figs/overview.tex
\begin{figure*} [t]
    \centering
    \includegraphics[width=\textwidth]{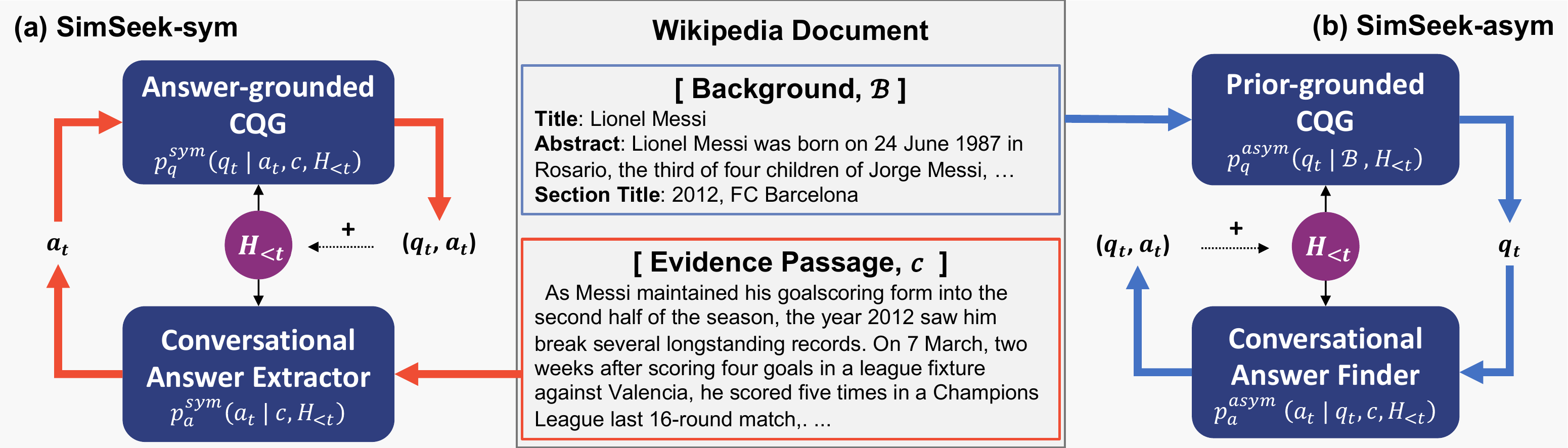}
    \caption{Overview of our frameworks continuing the held-out conversation $\mathcal{H}_{<t}$. To generate a QA pair $(q_t, a_t)$ at current turn $t$,  (a) \textsc{SimSeek-sym} first extracts the answer candidate $a_t$ from the passage $c$.
    Then, the questioner generates question $q_t$ that can be answered by $a_t$.
    % Both models can access the same evidence passage. 
    (b) \textsc{SimSeek-asym} first asks a follow-up question without accessing the passage.
    The answerer then provides an answer to the question from the evidence passage.
    Finally, we append the resulting QA pairs to the history $\mathcal{H}_{<t}$ for moving on to the next turn.
    % consists of prior-grounded CQG and conversational answer finder to simulate information-aymmetric conversation. Each model access background knowledge $\mathcal{B}$ and evidence passage $c$ to output question and answer, respectively.
    }
    \label{fig:overview}
\end{figure*}

%% file: tabs/03_method.tex
\section{SimSeek}
 We newly introduce two opposite ways to simulate synthetic conversations from unlabeled documents, \textsc{SimSeek-sym} and \textsc{SimSeek-asym}, as illustrated in Figure~\ref{fig:overview}.

\subsection{SimSeek-sym}
\label{sec:simseek-sym}
We propose a strong baseline, \textsc{SimSeek-sym}, inspired by the information-symmetric scenario. 
It can also be viewed as a straightforward extension of QA generation frameworks that are dominant in single-turn QA tasks~\cite{puri2020training, lewis2021paq}.
The framework is composed of the following components:
\begin{enumerate}
    \item A \textit{conversational answer extractor} (CAE) to detect answer candidates from the passage, considering the conversation.
    \item An \textit{answer-grounded CQG} (CQG$_{answer}$) to generate conversational questions that are likely to be answered by the detected candidates.
    \item A \textit{filtering} CQA model that predicts an answer to the generated question based on the conversation. 
    If the predicted answer is not matched with the predetermined answer by CAE, the QA pair is dropped. 
\end{enumerate}

\paragraph{Conversational Answer Extractor}
The component identifies spans that are likely to become an answer to the probable questions from the passage.
The selected span should also be natural to keep the conversational flow.
Specifically, the CAE model $p_a^{sym}(a_t \mid \mathcal{H}_{<t}, c)$ calculates the likelihood of answer span $a_t$ and predicts the most likely prediction $\hat{a}_t$ without taking the current question $q_t$.
By the likelihood values, we obtain the set of top-$k$ answer candidates $\hat{A}_t=\{\hat{a}^1_t, \hat{a}^2_t, \dots, \hat{a}^k_t\}$.
By jointly encoding the history $\mathcal{H}_{<t}$ with the passage $c$, the component could consider conversational flow when extracting $\hat{A}_t$.
We adapt 2D span extraction head upon the backbone architecture as \citet{lewis2021paq} propose.

\paragraph{Answer-grounded CQG}
Grounded on each extracted span, the CQG$_{answer}$ generates a follow-up question on the held-out conversation.
% Given answer spans extracted from it, the CQG$_{answer}$ generates a follow-up question on the held-out conversation.
Thus, it should satisfy multiple objectives at once; generating proper questions for the answer and coherent with the history.
% Thus, the resulting questions should not only fit into the given answer but also be coherent with the conversational context.
Formally speaking, the CQG$_{answer}$ synthesizes the conversational question based on the history, passage, and extracted answer, i.e. $p_{q}^{sym}(q_t \mid c, a_t, \mathcal{H}_{<t})$. 
We employ a T5-based sequence-to-sequence model as a backbone of the component~\cite{raffel2020exploring}. 
In particular, we highlight target answer $a_t$ as rationale span in the passage $c$ with a special token suggested by \citet{gu2021chaincqg}. 
In addition, we adopt a mask prediction scheme that aligns its objective with that of the pre-training phase following \citet{chada2021fewshotqa}.
% , shown to be sample efficient in prior work~\cite{chada2021fewshotqa}.
% since its objective is complementary with that of pre-training phase~\cite{chada2021fewshotqa}. 
% Performance of CQG models
% In our pilot experiments, our module achieves competitive performance with other current approaches in the CQG task.

\paragraph{Roundtrip Filtration for CQA}
The filtering model ensures the quality of generated questions, by checking the roundtrip consistency~\cite{alberti2019synthetic, puri2020training, lewis2021paq}. When the predictions of the filtering model are not matched with the predetermined answer by CAE, the question-answer pairs are discarded. 
We ease the filtering rule, from exact match to word-level similarity (i.e., F1 score) since the answers in CQA are often lengthy, compared to those in the single-turn QA. 
% Previous works~\cite{alberti2019synthetic, puri2020training, lewis2021paq} in QA generation have adopted the trained QA model as the filterer.
% and then discarded questions when predictions of filterers are not matched exactly with the grounded answer.
We employ the fine-tuned CQA model as our filtered, i.e., $p_f^{sym}(a_t \mid q_t, c, \mathcal{H}_{<t})$.
% Since answer texts to open-ended questions in CQA often longer, we ease the filtering rule, from exact match to thresholds by F1 score.

\subsection{SimSeek-asym}
 To simulate the information-asymmetric conversation effectively, we introduce a novel framework, \textsc{SimSeek-asym} (Figure~\ref{fig:overview} (b)).
 The framework consists of the following components: 
 \begin{enumerate}
     \item A \textit{prior-grounded CQG} (CQG$_{prior}$) for generating conversational questions relying solely on prior information (i.e., background information relevant to the topic).
     \item A \textit{conversational answer finder} (CAF) to comprehend the generated question and provides the most acceptable answer to the question from the evidence passage.
 \end{enumerate}

\paragraph{Prior-grounded CQG} CQG$_{prior}$ asks questions from insufficient information.
% to simulate the information-asymmetric conversation. 
Hence, the component requires neither the answer at the current turn nor the answer-containing passage.
Instead, it generates questions solely based on the background information about the topic, $\mathcal{B}$.
Specifically, it models conversational question $q_t$ from the given history $\mathcal{H}_{<t}$ and background $\mathcal{B}$, i.e. $p_{q}^{asym}(q_t \mid \mathcal{H}_{<t}, \mathcal{B})$. 

For a fair comparison of two CQG components, T5-based sequence generator is adopted to implement the CQG$_{prior}$, same with the CQG$_{answer}$.
They share the same architecture but their designs differ from each other.
% TODO: 더 많은 정보 answerCQG vs 부족한 정보 prior-CQG
We restrict CQG$_{prior}$ from accessing answer-relevant information, encouraging it to learn information-seeking behavior.
Although it slightly sacrifices QG performance in the automatic metric (i.e., BLEU), CQG$_{prior}$ plays a crucial role in simulating realistic information-seeking conversations.
We also demonstrate a one-to-one comparison by performing an intrinsic evaluation in Appendix~\ref{appendix:instrinsic}.

\paragraph{Conversational Answer Finder}
The conversational answer finder (CAF) provides an answer to the generated question based on the evidence passage.
Its objective is modeling $p_a^{asym}(a_t \mid q_t, c, \mathcal{H}_{<t})$. 
% Different from the CQG$_{prior}$ that does not access to the passage $c$, 
CAF plays the answerer's role in the information-seeking scenario by providing the requested information from the passage $c$. 
Note that any CQA model can be adopted as the CAF component and hence trained on the existing CQA datasets in the same way.
The design choice enables \textsc{SimSeek-asym} to generalize toward other advanced CQA approaches effectively.

\subsection{Synthetic CQA from Documents}
Our two \textsc{SimSeek} frameworks can generate synthetic conversations from unlabeled documents.
To train all modules in our frameworks, we use finite amount of human-labeled dataset $\mathcal{D} = \{(\mathcal{B}^i, c^i, \mathbf{q}^i, \mathbf{a}^i)\}_{i=0}^{|\mathcal{D}|}$, where the $\mathbf{q}^{i}$ and $\mathbf{a}^{i}$ denote all questions and answers, respectively, up to the maximum turn $T$ in $i$-th conversation.

In the inference phase, we suppose a unseen corpus $\mathcal{C} = \{(\mathcal{B}^j, c^j)\}_{j=0}^M$, which contains total $M$ number of unlabeled documents.
Each \textsc{SimSeek} framework sequentially synthesizes questions and answers at every turn $t$, starting from empty history $\mathcal{H}_{<0}$.
% TODO : temrination condition
% where $\mathcal{H}_0$ is a empty string so that the framework initializes a conversation only with $\mathcal{B}$ at the first turn.
Specifically, \textsc{SimSeek-sym} is formulated as:
\begin{equation*}
\label{eqn:simseek_sym}
\begin{aligned}
    & p^{sym}(q_t,a_t \mid \mathcal{H}_{<t}, \mathcal{B}, c) \\
    & \approx p^{sym}_q(q_t \mid a_t, \mathcal{H}_{<t}, c) \cdot p^{sym}_a(a_t \mid \mathcal{H}_{<t}, c) &
\end{aligned}
\end{equation*}
It first narrows down the potential target of the question, constraining the question distribution.
% TODO : limitation => , which hinders the questioner from ...
Note that it does not consider the filtering process while generating the conversation. 
Instead, we discard unqualified $(q, a)$ pairs after all conversations are terminated.
% Note that the filtering model does not terminate the conversation while two components (i.e., $p^{sym}_a$ and $p^{sym}_q$) sequentially generate questions or answers. 
% Instead, we obtain predictions from the filtering model and pick consistent pairs after whole conversations are generated.
On the other hand, \textsc{SimSeek-asym} decomposes the process into:
\begin{equation*}
\begin{aligned}
    & p^{asym}(q_t,a_t \mid \mathcal{H}_{<t}, \mathcal{B}, c) \\
    & \approx p^{asym}_a(a_t \mid q_t, \mathcal{H}_{<t}, c) \cdot p^{asym}_q(q_t \mid \mathcal{H}_{<t}, \mathcal{B})
\end{aligned}
\end{equation*}
Contrary to \textsc{SimSeek-sym}, it allows the question distribution to approximate any questions relevant to the topic. 
Finally, our frameworks generate question and answer at every turn $t$ as:
\begin{equation*}
    \hat{q}_{t}, \hat{a}_{t} = \operatorname*{arg\,max}_{q_t, a_t}~ p(q_t,a_t \mid \hat{\mathcal{H}}_{<t}, \mathcal{B}, c)
\end{equation*}

where $\hat{\mathcal{H}}_\text{<t}$ is a sequence of the generated $(q, a)$ pairs at previous turns and $p(\cdot)$ can be modeled as either $p^{sym}(\cdot)$ or $p^{asym}(\cdot)$.
The generated pair $(\hat{q}_t, \hat{a}_t)$ is appended to $\hat{\mathcal{H}}_\text{<t}$, resulting in  $\hat{\mathcal{H}}_\text{<(t+1)}$. The conversation progresses until it reaches the maximum turn $T$ or satisfies several termination rules
\footnote{~See each termination rule in Sec~\ref{subsec:ssl_setup} and \ref{subsec:wiki_setup}, respectively}.
Finally, we obtain sequences of the generated questions $\hat{\mathbf{q}}_j$ and answers $\hat{\mathbf{a}}_j$ by iterating the generation process, which results in the synthetic CQA dataset $\hat{\mathcal{D}} = \{(\mathcal{B}^j, c^j, \hat{\mathbf{q}}^j, \hat{\mathbf{a}}^j)\}_{j=0}^{M}$.

% the modules sequentially generates pseudo conversational questions $\hat{\mathbf{q}}$ and answers $\hat{\mathbf{a}}$ on $M$ number of unlabeled documents $\mathcal{C} = \{(\hat{\mathcal{B}_j}, \hat{c_j})\}_{j=0}^M$ and finally results $\hat{\mathcal{D}} = \{(\hat{\mathcal{B}}_j, \hat{c}_j, \hat{\mathbf{q}}_j, \hat{\mathbf{a}}_j)\}_{j=0}^{M}$.

%% file: tabs/04_experements.tex
\section{Evaluating Synthetic Conversations}

We evaluate our \textsc{SimSeek} in the semi-supervised setup.
To this end, we train all components on the existing CQA dataset $\mathcal{D}$ first. 
Then, synthetic conversations are generated upon unseen documents $\mathcal{C}$ by our frameworks.
% Assuming in-domain unlabeled documents are provided, we apply the described frameworks to generate synthetic conversations upon unlabeled documents.
% To evaluate synthetic conversations, we use them as additional training set for downstream tasks.
The resulting conversations are used as an additional training resource for downstream tasks. 
We train task-specific backbones on the synthetic datasets and test their performances in two downstream tasks, CQA and conversational search. 
See more details in Appendix~\ref{appendix:experimental_details},~\ref{appendix:implementation_details}.
% Then, we train task-specific models on synthetic datasets and compare their performances in two downstream tasks, CQA and conversational search.

% In this section, we first specify our experimental setup and baselines for generating synthetic conversations.
% % (\textsection~\ref{subsec:experimental_setup})
% We then demonstrate the effectiveness of our frameworks, comparing them to the baseline frameworks.

\subsection{Experimental Setup}
\label{subsec:experimental_setup}
\paragraph{Datasets}

All baselines and our frameworks are trained on a recent CQA benchmark, QuAC~\cite{choi2018quac}, which consists of 100k QA pairs for information-seeking conversation.
CANARD~\cite{elgohary2019can} convert questions in QuAC into self-contained questions such that they could be understood without the conversation.
We construct a single-turn QA dataset by replacing questions in QuAC with them, which is called CANARD in below.
For evaluating the quality of synthetic conversations in the conversational search, we use OR-QuAC~\cite{qu2020open}.
It extends QuAC to the open-domain setup and measures the performance in the passage retrieval task.
Further details are described in Appendix~\ref{appendix:datasets_detail}.

\paragraph{Semi-supervised Setup}
\label{subsec:ssl_setup}
We split the original training set of QuAC into three subsets, QuAC$_{seen}$, QuAC$_{unseen}$, and the validation set, following prior works~\cite{elgohary2019can,qu2020open}\footnote{~Table \ref{table:data_stat} shows detailed statistic}.
We train all components on QuAC$_{seen}$, considering it as the given dataset $\mathcal{D}$.
Then, assuming unlabeled documents of QuAC$_{unseen}$ as an unseen corpus $\mathcal{C}$, we construct synthetic dataset $\hat{\mathcal{D}}$ with the CQA generation frameworks.

We use synthetic datasets detailed in Section ~\ref{sec:anal_datasets}.
In particular, we set the maximum turn T as 6 and do not end the conversation early than that.
Models for downstream tasks are trained on either the synthetic set only ($\hat{\mathcal{D}}$) or the merged set ($\mathcal{D} + \hat{\mathcal{D}}$), and tested on the original evaluation set.
We report and compare their performances to measure the quality of synthetic conversations.
More details are in Appendix~\ref{appendix:ssl_detail}

\paragraph{Baselines for Synthetic CQA Generation}

We introduce solid baselines for synthesizing CQA datasets and compare them with our methods.
Since there doesn't exist any prior work available, we simply extend PAQ~\cite{lewis2021paq}, one of the dominant frameworks in single-turn QA tasks, and then adopt them as our baselines. 
For PAQ-CANARD baseline, we train its all components on CANARD, regarding it as the single-turn QA task.
PAQ-QuAC extends it by adopting CQG$_{answer}$ that can consider the conversation history.
The questioners in all baselines also require the target answer when generating questions.
But all answerers cannot consider the held-out conversation.
More details are in Appendix~\ref{appendix:baseline_generaiton}

\paragraph{Baselines in Downstream Tasks}
After building synthetic conversations, we train and test the baseline models in downstream tasks.
For the CQA task, we choose three backbone architectures, RoBERTa \cite{liu2019roberta} in base and large size, and Longformer-large \cite{Beltagy2020Longformer}.
Longformer architecture has been shown to be effective for encoding much longer history, which achieves competitive performance with the previous state-of-the-art approach~\cite{zhao2021ror}.
In addition, we test synthetic datasets in one of the document retrieval tasks, conversational search, where systems are required to retrieve relevant documents to conversational queries.
We employ the off-the-shelf dense retriever, DPR~\cite{karpukhin2020dense}, as our baseline for conversational search.
Further details are in Appendix~\ref{appendix:baseline_downstream}

\subsection{Experimental Results}

\input{tables/unseen}

% We demonstrate the effectiveness of our frameworks, comparing them to the aforementioned baseline approaches.
% In the following experiments, every approach is measured with the end performance in downstream tasks.
% Specifically, we train backbone models on the generated dataset and evaluate them on each benchmark.

\paragraph{Semi-supervised CQA}
\label{subsec:in_domain_experiment}

We evaluate the effectiveness of our synthetic datasets on a recent CQA benchmark, QuAC.
Table~\ref{table:semi_quac} shows the end performance of CQA models trained on the resulting datasets.
% which leverages both labeled ($\mathcal{D}$) and synthetic ($\hat{\mathcal{D}}$) datasets. 
% ``Human Annot.'' indicates the CQA models are trained on the human-labeled dataset, i.e., QuAC$_{unseen}$; thus we consider them as an upper bound of synthetic dataset generation. 
% On the other hand, the ``None'' is a de-facto baseline that does not use any unlabeled corpus $\mathcal{C}$, but rather trained on the labeled dataset $\mathcal{D}$ and its perturbed dataset, i.e., $|\hat{\mathcal{D}}| = 0$.
% Similarly, Back-Translation and De-contextualization do not take the $\mathcal{C}$ but only perturb or transform the questions in the $\mathcal{D}$. The question augmentation methods often degrade the baseline performance (None), showing limited improvement only in the RoBERTa-large backbone.
% Note that they are unavailable to generate the synthetic dataset $\hat{\mathcal{D}}$ from the unlabeled corpus since they require the original question-answer pairs.
% Baselines
 When using the synthetic dataset alone ($\hat{\mathcal{D}}$), PAQ-CANARD shows the lowest performance in all CQA backbones.
 It implies the difficulty of directly extending single-turn QA methods to CQA. 
Adopting CQG module that can consider the conversational context (PAQ-QuAC) advances CQA performance with a huge gap of over 15 F1 scores in all backbones.
It indicates that learning to comprehend conversational questions is crucial to improving CQA performance.
\textsc{SimSeek-sym}, where all components consider the history, shows comparable scores with PAQ-QuAC.
Despite the small gap, \textsc{SimSeek-sym} largely outperforms PAQ-QuAC without the filtering process that often saturates the end-CQA performances.
\footnote{~Table~\ref{table:semi_nofilter} provides an ablation study on it}
\textsc{SimSeek-asym} shows dominant performance compared to other baselines.
Moreover, the performance gap increases as the size of CQA models gets larger.
It implies that \textsc{SimSeek-asym} could generate finer quality of synthetic conversations when leveraging better CQA models.

Results of ($\mathcal{D}$ + $\hat{\mathcal{D}}$) show augmentation effect of generated datasets.
Most of the baselines fail to improve the performance, which implies the difficulty of generating realistic CQA examples.
% We also conduct a qualitative analysis on why PAQ-QuAC and SimSeek-sym perform poorer than PAQ-CANARD (see Section [ref].)
% Ours
On the other hand, our proposed framework \textsc{SimSeek-asym} consistently improves CQA performance over all CQA backbones.
Specifically, it improves RoBERTa-large and Longformer-large by 1.9 and 1.1 F1 scores compared to the main baseline (None), respectively. 
% In particular, improvement is more significant in large size of models.
Surprisingly, Longformer-large with \textsc{SimSeek-asym} achieves competitive performance as when trained on the human-labeled dataset, by a gap of only 0.7.
It shows \textsc{SimSeek-asym} succeeds in simulating human-like conversations.
% Longformer-large achieves competitive performance to humans with a gap of only 0.7 when using \textsc{SimSeek-asym}.

% Longformer-large trained on synthetic dataset by DAC-asym achieves competitive performance to that on humans with a gap of only 0.7.

% While the performances of other baseline methods are  regardless of the CQA backbone, SimSeek-asym achieves a better score with a more capable backbone.

\input{tables/retrieval}

% \subsection{Large-scale Experiment on Wikipedia}
% \input{tables/wiki}
% In Table \ref{table:full_quac}, we report the performance improvements when leveraging SimSeek-asym upon Wikipedia corpus.

\paragraph{Utility in Conversational Search}

Table \ref{table:orquac} shows the retrieval performances of baseline retrieval model, DPR \cite{karpukhin2020dense} on OR-QuAC dataset~\cite{qu2020open}.
% Training the DPR with more steps without additional synthetic dataset significantly degrades the performance, which implies overfitting.
% SimSeek-sym performs better than baseline but still fails to lift the score.
The resulting conversations from \textsc{SimSeek-sym} degrade the retrieval performance, indicating it fails to model questions in the information-seeking conversation.
We report the results of \textsc{SimSeek-asym} combined with different answerers.
% Improvement from data augmentation is greater as the better answerer is involved in simulating the conversation.
Among them, the framework with Longformer-large only succeeds in boosting the retrieval performance of DPR.
Although dense retrievers do not encode any answers by the task setup, retrieval performances vary depending on the capabilities of answerer model.
It implies that interacting with a better answerer allows the questioner to ask more diverse and adequate questions, leading to a more realistic information-seeking conversation.
% that are likely to be asked in information-asymmetric conversation.

%% file: tables/unseen.tex
\begin{table}[t]
    \small
    \centering
    \begin{threeparttable}
    \begin{tabular*}{0.98\columnwidth}{l|cc}
        \toprule
        \multicolumn{1}{l|}{\multirow{1}{*}{\textbf{CQA Backbone}}} & \multicolumn{2}{c}{\multirow{1}{*}{Trained on}} \\
        % & & & \\
        \multicolumn{1}{l|}{\quad Synthetic CQA Generation} & $\hat{\mathcal{D}}$ & $\mathcal{D}$ + $\hat{\mathcal{D}}$  \\
        \midrule
        \midrule
        \textbf{RoBERTa-base} & &   \\
        \quad None ($\hat{\mathcal{D}} = $ empty) & - & 64.4 \\
        \quad PAQ-CANARD & 38.2 & 64.3 \\
        \quad PAQ-QuAC & 55.9 & 64.6 \\
        \quad \textsc{SimSeek-sym} & 55.5 & 64.4 \\
        \quad \textsc{SimSeek-asym} & \textbf{62.5} & \textbf{65.3} \\
        \midrule
        \quad Human Annot. ($\hat{\mathcal{D}} = $ QuAC$_\text{unseen}$)  & 65.3 & 67.5 \\
        % & + I-RTC & 62.7 & 65.1 \\
        \midrule \midrule
        % \multicolumn{1}{c}{\textbf{RoBERTa-large} trained on} & $\hat{\mathcal{D}}$ & $\mathcal{D}$ + $\hat{\mathcal{D}}$  \\
        % \midrule
        \textbf{RoBERTa-large} & &   \\
        \quad None ($\hat{\mathcal{D}} = $ empty) & - & 65.6 \\
        \quad PAQ-CANARD & 38.8 & 66.5 \\
        \quad PAQ-QuAC & 51.5 & 66.6 \\
        \quad \textsc{SimSeek-sym} & 54.3 & 66.3 \\
        \quad \textsc{SimSeek-asym} & \textbf{64.8} & \textbf{67.5} \\
        \midrule
        \quad Human Annot. ($\hat{\mathcal{D}} = $ QuAC$_\text{unseen}$) & 65.0 & 70.3 \\
        % & + I-RTC & \textbf{64.9}  & 67.6 \\
        \midrule \midrule
        % \multicolumn{1}{c}{\textbf{Longformer-large} trained on} & $\hat{\mathcal{D}}$ & $\mathcal{D}$ + $\hat{\mathcal{D}}$  \\
        % \midrule
        \textbf{Longformer-large} & &   \\
        \quad None ($\hat{\mathcal{D}} = $ empty) & - & 72.0 \\
        \quad PAQ-CANARD & 37.5 & 71.5 \\
        \quad PAQ-QuAC & 61.7 & 71.7 \\
        \quad \textsc{SimSeek-sym}  & 60.8 & 71.7 \\
        \quad \textsc{SimSeek-asym} & \textbf{71.5} & \textbf{73.1} \\
        \midrule
        \quad Human Annot. ($\hat{\mathcal{D}} = $ QuAC$_\text{unseen}$) & 72.3 & 73.8 \\
        % & + I-RTC & 71.1 & - \\
        \bottomrule
    \end{tabular*}
    \end{threeparttable}
    \caption{Comparison over synthetic CQA generation methods. We report F1 scores for the end CQA performance on the development set of QuAC. Frameworks for synthetic CQA generation are trained on the original dataset $\mathcal{D}$ and generate the snythetic dataset $\hat{\mathcal{D}}$. Finally, student CQA baselines are fine-tuned on either $\hat{\mathcal{D}}$ or $\mathcal{D} + \hat{\mathcal{D}}$. ``Human Annot.'' indicates human-labeled conversations from the original QuAC, i.e., QuAC$_\text{unseen}$.}
    \label{table:semi_quac}
\end{table}

%% file: tables/retrieval.tex
\begin{table}[t!]
    \small
    \centering
    \begin{threeparttable}
    \setlength{\tabcolsep}{0.5em}
    \begin{tabular*}{0.85\columnwidth}{lccc}
        \toprule
        \textbf{Retrieval Model} & \multicolumn{3}{c}{OR-QuAC} \\
        \quad Synthetic CQA & MRR & R@5 & R@20 \\
        \midrule
        \multicolumn{4}{l}{\textbf{DPR} trained on \textit{$\mathcal{D}$ + $\hat{\mathcal{D}}$}}  \\
        \hspace{0.5em} None ($\hat{\mathcal{D}}$ = empty) & 53.3 & 64.8 & 73.8 \\
        % \hspace{0.5em} \hspace{0.3em} w/ further training & 48.2 & 60.5 & 72.1 \\
        \hspace{0.5em} \textsc{SimSeek-sym} & 50.4 & 62.4 & 72.3 \\
        \hspace{0.5em} \textsc{SimSeek-asym} & &  &  \\
        \hspace{0.5em} \hspace{0.3em} w/ RoBERTa-base & 51.5 & 63.3 & 73.6 \\
        \hspace{0.5em} \hspace{0.3em} w/ RoBERTa-large & 53.4 & 64.4 & 73.6 \\
        \hspace{0.5em} \hspace{0.3em} w/ Longformer-large & \textbf{54.4} & \textbf{66.1} & \textbf{75.3} \\
        \bottomrule
    \end{tabular*}
    \end{threeparttable}
    \caption{Evaluation results of conversational passage retrieval on OR-QuAC test set. Longformer-large architecture is used for \textsc{SimSeek-asym}.}
    \label{table:orquac}
\end{table}

%% file: tabs/05_analysis.tex
\section{Analysis}
\label{sec:anal_datasets}

We report detailed statistics of the generated datasets and perform a human evaluation to analyze the quality of conversations and compare them.

\subsection{Qualitative Analysis}
Table~\ref{table:stat_comparison} summarizes statistics of synthetic conversations from two frameworks. All datasets show similar overlap score of question-answer, word-level F1 of ($q_t$, $a_t$).
In contrast, we observe a meaningful gap in the overlap between the question and previous responses, word-level F1 of ($q_t$, $a_{0:(t-1)}$), which measures how many words from the opponent's responses are reused in the current question.
% We will conjecture the difference in following subsection.
% In question-answer overlap, no significant differences are observed between the synthetic datasets.
On the other hand, \textsc{SimSeek-sym} more frequently exploits one of the tricks for seeking new information, ``Anything else?'' questions\footnote{~See the example in Table~\ref{table:case_study}.}, which shifts the current topic and requests any new information.
The questioners ask these questions effortlessly without considering conversational context much.
\textsc{SimSeek-sym} rarely asks unanswerable questions. On the other hand, \textsc{SimSeek-asym} often fails to acquire answers as humans do and their frequencies are similar.
\textsc{SimSeek-asym} generates conversations that have similar statistics to the original QuAC, overall.

\input{tables/stat_comparison}
\subsection{Human Evaluation}
We perform a human evaluation to examine the quality of synthetic conversations. Specifically, we conduct a pairwise judgment with Amazon Mechanical Turk, asking the workers to assess the relative quality of follow-up QA pairs. More details are described in Appendix~\ref{app:mturk}.

We first ask the workers to judge (1) the overall adequacy of generated QA pairs to the given history.
It represents how adequate the QA pair is for continuing the given conversation.
Additionally, we also ask (2) informativeness (i.e., does the question try to gather new information), (3) context relevance (i.e., how relevant or specific is the question to the given context), and (4) answer accuracy (i.e., whether the answer is a correct one to the question), inspired by the metrics of \citet{qi2020stay, li2021ditch, thoppilan2022lamda}. 
Figure~\ref{fig:human_eval} summarizes the results. We find that there are no significant differences in informativeness. Hence, \textsc{SimSeek-asym} show similar informativeness scores compared to humans.
% As a result, no significant differences are observed in informativeness.
% Conversations from \textsc{SimSeek-asym} show similar informativeness score, compared to humans.
% It implies that our frameworks generate questions as informative as humans.

\textbf{\textsc{SimSeek-sym} asks questions closely related to context and answer; however they are rarely adequate.}
The annotators conclude that conversations from \textsc{SimSeek-sym} are more relevant to the document than \textsc{SimSeek-asym}. Moreover, answers to their corresponding questions are more accurate than \textsc{SimSeek-asym} since the questioner asks questions with having their answers.
However, \textsc{SimSeek-sym} are less frequently chosen as adequate.
In other words, it succeeds in generating coherent QA pairs at each turn, but it is inadequate in simulating information-seeking behaviors.
% move the conversation toward meaningful direction.

\textbf{Conversations from \textsc{SimSeek-asym} are reviewed as more adequate even than humans, overall.}
We observe that repeating the opponents' responses often leads to high adequacy\footnote{~We further report the qualitative case study in Table~\ref{table:case_study}}.
As shown in Table~\ref{table:stat_comparison}, \textsc{SimSeek-asym} asks questions highly overlapped with previous responses $a_{0:(t-1)}$, which makes it perceived as more helpful and communicative by the annotators.
On the contrary, original questions in QuAC often request new knowledge concisely, which seems relatively less enthusiastic.
Thus, this leads to the ironic result that human-annotated conversations are less frequently chosen in terms of overall adequacy.

\input{figs/human_analysis}

%% file: tables/stat_comparison.tex
\begin{table}[t!]
    \small
    \centering
    \begin{threeparttable}
    \begin{tabular*}{0.90\columnwidth}{lrrr}
        \toprule
         & \multirow{2}{*}{QuAC} & \multicolumn{2}{c}{\textsc{SimSeek}} \\
         & & \textsc{sym} & \textsc{asym} \\
        \midrule
        tokens / question & 6.5 & 7.3 & 7.5 \\
        tokens / answer & 15.1 & 17.6 & 16.8 \\
        \midrule
        % \% $(q,a)$ overlap & 4.1 & 5.9 & 5.7 \\
        F1 of $(q_t, a_t)$ & 7.0 & 10.2 & 9.7 \\
        F1 of $(q_t, a_{0:(t-1)})$ & 17.1 & 19.0 & 26.9 \\
        \midrule
        \% Anything else? & 18.4 & 23.8 & 17.0 \\
        \% Unanswerable Qs & 17.3 & 1.0 & 19.7 \\
        % \midrule
        % \% Frequent Qs & 27.9 & 28.3 & 21.8 \\
        % Informativeness & 12.8 & 16.7 & 15.8 \\
        % Specificity & 65.0 & 59.1 & 66.4 \\
        % Informativeness & 71.5 & 83.8 & 66.7 \\
        % Answer Relevance & 89.0 & 97.0 & 87.3 \\
        \bottomrule
    \end{tabular*}
    \end{threeparttable}
    \caption{Comparison over the original QuAC and synthetic datasets from our frameworks. 
    \textsc{SimSeek-asym} uses Longformer-large as the answerer in the table. 
    For scalable analysis, we automatically count ``Anything else?'' questions with certain strings (e.g., ``other'' and ``else'')}
    \label{table:stat_comparison}
\end{table}

%% file: figs/human_analysis.tex
\begin{figure} [t!]
    \centering
    \includegraphics[width=\columnwidth]{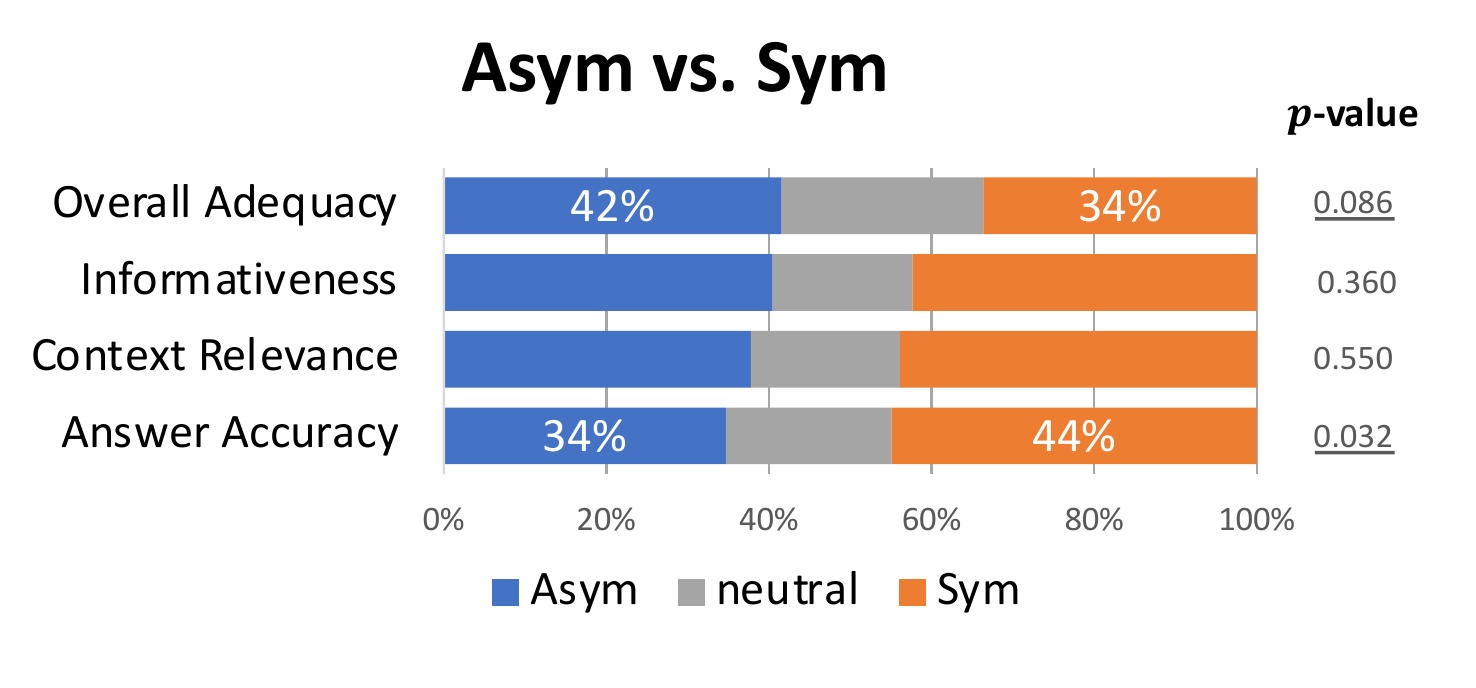}
    \includegraphics[width=\columnwidth]{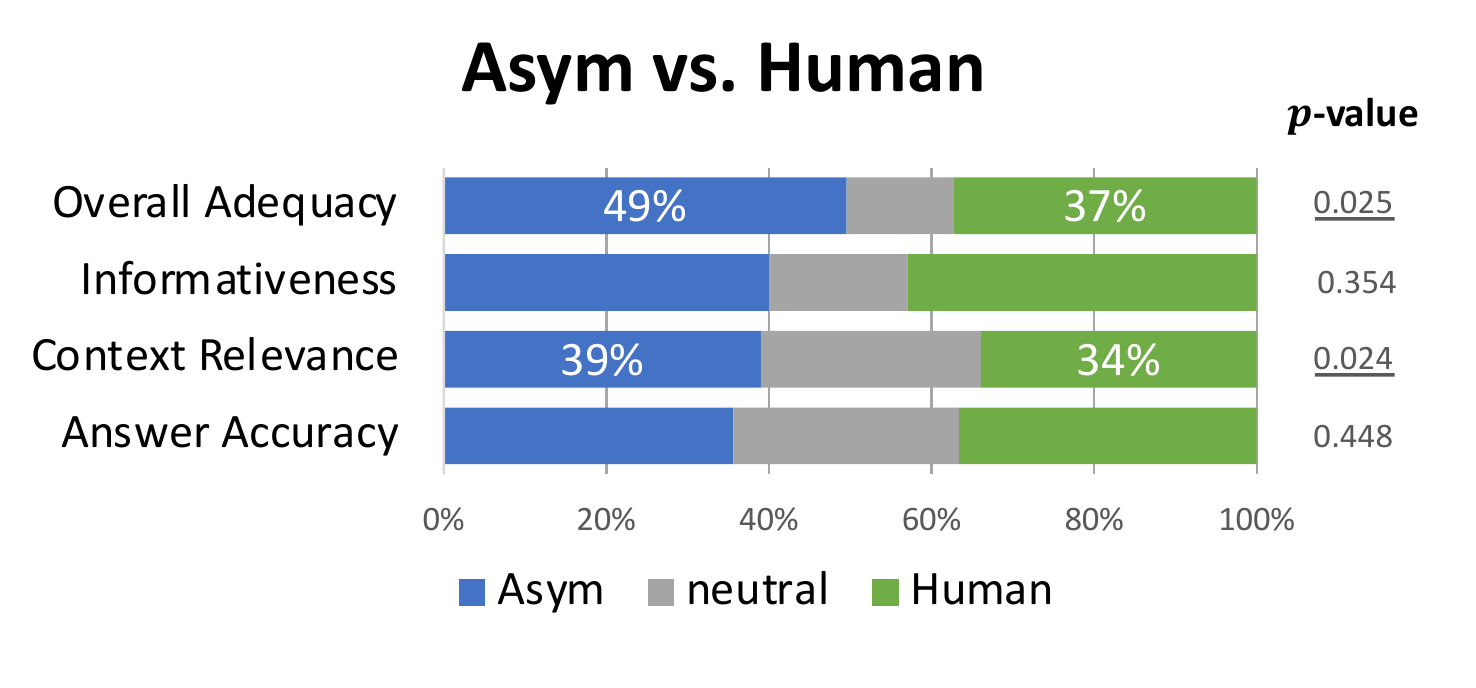}
    \caption{Pairwise human evaluation results for the original QuAC (human) and the synthetic conversations from \textsc{SimSeek} (sym and asym). 
    The annotators pick the better one from two random QA pairs in terms of each criterion.
    We report the overall proportion of majority votes for each instance.
    We conduct a bootstrap test with $10^5$ samples for the difference between pairs. 
    We add labels only when the difference is statistically significant ($p$-value lower than 0.1).}
    \label{fig:human_eval}
\end{figure}

%% file: tabs/06_wiki.tex
\section{Wiki-SimSeek}
Based on our results, we newly construct synthetic conversations on the larger scale of corpus, Wikipedia, by applying \textsc{SimSeek-asym}.
We finally release a large-scale resource of information-seeking conversations, \textsc{Wiki-SimSeek}, which consists of 2.1 million questions and answer pairs upon 213k Wikipedia passages.

\subsection{Dataset Construction}
\label{subsec:wiki_setup}
\input{tables/data_comparison}
First, we crawl the documents from Wikipedia by using KILT tools~\cite{petroni2021kilt} \footnote{~\href{https://github.com/facebookresearch/KILT}{github.com/facebookresearch/KILT}}.
Following \citet{choi2018quac}, we collect Wikipedia articles from a list of category keywords leveraging a web interface, Wikipedia foundation\footnote{~\href{https://petscan.wmflabs.org/}{petscan.wmflabs.org}}.
We use the abstract of Wikipedia documents as background information, while sections between 250 and 550 words are picked as evidence documents. 
Then, we use \textsc{SimSeek-asym} with the answerer (Longformer-large) to simulate synthetic conversations upon the crawled Wikipedia documents.
In particular, \textsc{SimSeek-asym} generates the conversations until they reach the twelfth turn or more than three unanswerable questions are asked. Table~\ref{table:data_comparison} shows overall statistics of \textsc{Wiki-SimSeek} and compares it to other QA datasets.
Each dialog contains 10.0 question-answer pairs on average, which shows that the framework can carry on long conversations and bring out new information.

\subsection{Further Improvement of CQA Models}
\input{tables/wiki}
To investigate the effect of \textsc{Wiki-SimSeek}, we compare our models with previous approaches on QuAC as shown in Table~\ref{table:full_quac}.
% Table~\ref{table:full_quac} compares the previous works in QuAC with our models.
Unlike the semi-supervised setup, all components of our frameworks are trained on the training set of QuAC. 
Further training Longformer-large on \textsc{Wiki-SimSeek} improves the performance by 1.0 of F1 and 1.5 of HEQ-Q.
To compare fairly with the previous best performing model RoR~\cite{zhao2021ror}\footnote{~\citet{zhao2021ror} also use CoQA to achieve the scores.}, we also employ another CQA dataset, CoQA~\cite{reddy2019coqa} as an additional training resource. 
Baseline performances are significantly degraded when we use only CoQA for data augmentation.
It shows that simply combining two different datasets could cause a distribution shift, leading to a performance drop. 
However, when \textsc{Wiki-SimSeek} is additionally used, it boosts the CQA model performance by a large gap of 1.7 F1 score.
It achieves state-of-the-art performance in F1 score on QuAC.
As a result, our dataset could be one of the solutions to mitigate the shift, providing further improvement.
% Note that authors of RoR also report they use CoQA as an additional training dataset~\cite{zhao2021ror}.
Note that other baseline approaches could be further improved by our synthetic datasets.
Moreover, since \textsc{SimSeek-asym} is a model-agnostic framework, other CQA models can be adopted as the answerer for generating synthetic datasets. 

% The promising results shows that it can be effective when using with other datasets.

% \subsection{Future Usage}

% \citet{choi2018quac} describe the limitations of QuAC dataset in their datasheet\footnote{\href{https://quac.ai/datasheet.pdf}{quac.ai/datasheet.pdf}}. 
% QuAC might contain bias over famous people since articles were filtered on the ``people'' category associated with other subcategories of Wikipedia. 
% Thus, the authors discourage model deployment trained on QuAC in real-world settings. 
% However, we hope our \textsc{Wiki-SimSeek} could be used for CQA models to generalize the category beyond ``people'' in the real scenario. Table~\ref{table:case_study_ood} shows a qualitative example in \textsc{Wiki-SimSeek}. 
% Surprisingly, \textsc{SimSeek-asym} successfully simulates conversations of unseen topics as well, even if it is trained on QuAC. Our dataset could be used for adapting unseen domains in CQA research.

%% file: tables/data_comparison.tex
\begin{table}[t!]
    \small
    \centering
    \resizebox{0.98\columnwidth}{!}{
    \begin{tabular}{lccc}
        \toprule
        
        Dataset & Domain & Dialogs & Ques. \\
        \midrule \midrule
        \multicolumn{4}{l}{\textbf{Single-turn QA}} \\
        \quad SQuAD & Wikipedia & & 107K  \\
        \quad Natural Questions & Wikipedia & & 307K \\
        \quad PAQ & Wikipedia & & 65M  \\
        \midrule
        \multicolumn{4}{l}{\textbf{CQA}}\\
        \quad QuAC & Wikipedia (People) & 13K & 98K \\
        \quad CoQA & 7 sub-domains & 13K & 127K \\
        \quad DoQA & Stack Exchange & 2K & 10K \\
        \midrule
        \multicolumn{4}{l}{\textbf{Open-Domain CQA}} \\
        \quad OR-QuAC & Wikipedia (People) & 5.6K & 40.5K \\
        \quad QReCC & Wikipedia & 14K & 81K \\
        \quad TopiOCQA & Wikipedia & 4K & 50K \\
        \midrule
        \textbf{Ours} & & & \\
        \quad \textsc{Wiki-SimSeek} & Wikipedia & 213K & 2.1M \\
        \bottomrule
    \end{tabular}}
    \caption{Comparison over CQA datasets with \textsc{Wiki-SimSeek}}
    \label{table:data_comparison}
    \vspace{-3mm}
\end{table}

%% file: tables/wiki.tex
\begin{table}[t!]
    \small
    \centering
    \resizebox{0.98\columnwidth}{!}{
    \begin{tabular}{lccc}
        \toprule
        \multirow{2}{*}{CQA Model} & \multicolumn{3}{c}{QuAC} \\
         & F1 & HEQ-Q & HEQ-D \\
        \midrule
        % BERT-base & 63.2 \\
        HAE \cite{qu2019bert} & 63.1 & 58.6 & 6.0 \\
        GraphFlow \cite{chen2019graphflow} & 64.9 & - & -\\
        HAM \cite{qu2019attentive} & 66.7 & 63.3 & 9.5 \\
        ExCorD \cite{kim2021learn} & 67.7  & 64.0  & 9.3  \\
        RoR \cite{zhao2021ror}$^{*}$ & 75.7 & \textbf{73.4} & \textbf{17.8}  \\
        \midrule
        \textit{Ours} \\
        Longformer-large & 74.0 & 71.0 & 13.7 \\
        \quad  + \textsc{Wiki-SimSeek} & 75.0 & 72.5 & 13.2  \\
        \quad  + CoQA & 69.5 & 63.3 & 7.7 \\
        \quad  + CoQA + \textsc{Wiki-SimSeek} & \textbf{76.1} & \textbf{73.4} & 16.4  \\
        \bottomrule
    \end{tabular}}
    \caption{Comparsion over baseline CQA models  on the development set of QuAC. 
    By using \textsc{SimSeek-asym}, \textsc{Wiki-SimSeek} is generated from unlabeled documents of Wikipedia . Models with asterisk ($^{*}$) are additionally trained on CoQA datatset. GraphFlow~\cite{chen2019graphflow} does not report HEQ scores.}
    \label{table:full_quac}
    \vspace{-3mm}
\end{table}

%% file: tabs/07_related_works.tex
\section{Related work}

\paragraph{Conversational Question Answering}

With the advent of recent large-scale CQA datasets \cite{choi2018quac, reddy2019coqa}, numerous studies proposed methods to resolve the challenging task.
Most works focused on developing model structures \cite{zhu2018sdnet, qu2019bert, qu2019attentive} that are specialized in the CQA task.
Several works demonstrated the effectiveness of the flow mechanism in CQA \cite{huang2018flowqa, chen2019graphflow}.
Most recently, leveraging self-contained questions \cite{kim2021learn} or encoding longer context \cite{zhao2021ror} have been shown to be effective in the task.

% retrieval
% Recent studies extend the task into open-domain setup.
% After the release of a dataset consisting of self-contained questions rewritten by human annotators~\cite{elgohary2019can}, open-domain CQA and conversational search tasks have gained attention ~\cite{dalton2019cast, anantha2021open, adlakha2021topiocqa}.
% Several studies feed self-contained questions to existing search engines such as BM25~\cite{yu2020fewshot, voskarides2020quretec, lin2021multistage, kumar2020making}.
% Another line of studies \cite{yu2021few, lin2021contextualized, kim2022saving} attempt to directly train dense retrievers to jointly encode the conversation and question.

\paragraph{Synthetic QA Generation}
Many question generation (QG) researches have sparked advances in various QA tasks~\cite{dhingra2018simple, dong2019unified, lewis2019unsupervised, alberti2019synthetic, puri2020training, lewis2021paq}. 
Most of early studies propose to generate them in a cloze-style \cite{dhingra2018simple, lewis2019unsupervised} or by using pre-defined templates \cite{fabbri2020template}.
Recent studies for the synthetic QA generation propose the pipeline strategies composed of three sub-phases, answer extraction, question generation, and various filtering steps such as round-trip filtration \cite{alberti2019synthetic, puri2020training, lewis2021paq}.
% The first step, answer extraction, identifies spans in the given passage that are likely to be answer to question.
% An off-the-shelf tools for Named Entity Recognition (NER) or the trained answer extraction models are adopted for extracting answer candidates.
% % TODO: PLM 미리 약어쓰기
% For the next step, question generation, neural models for text generation are trained to generate coherent questions with the given answer and passage.
% % TODO: Answer-prior 를 활용하는 점 강조 
% Lastly, after generating synthetic $(q,a)$ pairs, a filtering QA model provides the predicted answer $\hat{a}$ given the generated question $q$.
% If the initial answer is not matched with the predicted one, the pair is discarded to ensure the round-trip consistency.

% These strategies have achieved remarkable success such that QA models trained on synthetic sets outperform them on the original training set.
% However, extending them to conversational setup is non-trivial.

\paragraph{Conversational Question Generation}

Many works attempt to generate human-like conversational questions. \citet{pan2019reinforced, gao2019interconnected} introduce the challenge of CQG and successfully extend the single-turn question generation to consider conversational input. 
% Their proposed models show effectiveness compared to single-turn question generators.
Most prior works are based on the information-symmetric assumption~\cite{pan2019reinforced, gao2019interconnected, nakanishi-etal-2019-towards, gu2021chaincqg}. Recently, \citet{qi2020stay} investigate information-asymmetric conversations. 
They first attempt to generate the conversational questions without evidence document.
Concurrently, \citet{dai2022dialog} propose a method to turn document into dialogue and release a large-scale dataset of synthetic dialogue.
However, they report improvements only in the conversational search task.

%% file: tabs/08_conclusion.tex
\section{Conclusion}

In this work, we propose a novel framework, \textsc{SimSeek}, simulating information-seeking conversation from given unlabeled documents. 
Our frameworks assume two scenarios and compare them to provide a deeper understanding of information-seeking conversation. Experimental result shows that our \textsc{SimSeek-asym} generates human-like conversation. Moreover, we provide insightful analyses to help understand the information-seeking conversation better. 
We finally release the large-scale resources of synthetic conversations, \textsc{Wiki-SimSeek}.
We hope it could be a stepping stone for building robust CQA models that can be generalized toward the real-world scenario.
Furthermore, it could be beneficial to identify the factors in realistic information-seeking conversations.

% As a result, we find the clue that asking questions with proper specificity and ambiguity is an important step to simulate more realistic conversations in terms of information-seeking.

%% file: tabs/99_limitations.tex
\section*{Limitations}
We tested various methods for automatically filtering the generated conversations from \textsc{SimSeek-asym}.
However, since it already simulates human-like questions as shown in our evaluation, we failed to significantly improve the performance in downstream tasks.
Thus, we only adopt several filtering rules to discard deviated conversations.
One might propose a novel filtration method for \textsc{SimSeek-asym} by investigating our resulting dataset, \textsc{Wiki-SimSeek}.
\textsc{Wiki-SimSeek}, is limited in the specific language  (i.e., English). 
We use machines with 8 V100 GPUs and training Longformer-large on \textsc{Wiki-SimSeek} takes a few days, which requires relatively high computational costs.

%% file: tabs/acknowledgement.tex
\section*{Acknowledgements}
We would like to appreciate Jinhyuk Lee, Hwanhee Lee, Miyoung Ko, Hyunjae Kim, Jaehyo Yoo, Hwaran Lee, Jungwoo Ha, and anonymous reviewers for providing constructive feedback.
This research was supported by the MSIT (Ministry of Science and ICT), Korea, under the ICT Creative Consilience program (IITP-2022-2020-0-01819) and the High-Potential Individuals Global Training Program (RS-2022-00155958) supervised by the IITP(Institute for Information \& communications Technology Planning \& Evaluation).
This work was funded by the National Research Foundation of Korea (NRF-2020R1A2C3010638).
This work was supported by the Hyundai Motor Chung Mong-Koo Foundation.

%% file: tabs/09_appendix.tex
\appendix

\section{Experimental Details}
\label{appendix:experimental_details}

\subsection{Datasets}
\label{appendix:datasets_detail}
\paragraph{QuAC}

QuAC \cite{choi2018quac} consists of 100k QA pairs in information-asymmetric dialogues, where a questioner asks questions based on a topic with background information, and an answerer returns the answers in the form of text spans in Wikipedia document. 
Restricting the questioners from accessing the answer-containing document, the authors encourage them to seek new information on a topic via conversation. 
Following \citet{choi2018quac}, we evaluate models with the F1 score for QuAC.
Since the test set is only available in the QuAC leaderboard, we evaluate models on the development set\footnote{~\href{https://quac.ai/}{quac.ai}}.
HEQ measures whether a CQA model finds more accurate answers than humans in each granularity (HEQ-Q for question, HEQ-D for dialogue)\footnote{~Evaluation scripts are provided by \href{https://quac.ai/}{quac.ai}}.

\paragraph{OR-QuAC}

\citet{qu2020open} extend the original QuAC dataset to open-domain setup\footnote{~\href{https://github.com/prdwb/orconvqa-release}{github.com/prdwb/orconvqa-release}}. It assumes that a ground-truth document is not given in advance, which means the answerers do not know what to be asked before a conversation begins. 
Instead, they first need to search relevant passages from web-scale documents (about 11M chunked passages) based on the given conversational history and current question.
% Then, the answer to the input can be predicted based on the retrieved passages. 
After reading the retrieved passage, they predict an answer to the question.
Following the original setup in \citet{qu2020open}, we only regard previous questions $\{q_1, q_2, ..., q_{t-1}\}$ as history without answers. 
% Because of the setup, we can compare qualities of questions from our synthetic datasets aside from answers on OR-QuAC.
% % Hence, if backbone architecture is controlled, we can test how the given questions are relevant to the passage by evaluating the resulting models.
% we can compare qualities of questions from our synthetic datasets aside from answers on OR-QuAC.
Since OR-QuAC is similarly partitioned with our QuAC splits (see details in Table~\ref{table:data_stat}), we use same synthetic conversations that are used for CQA task.
For evaluation, mean reciprocal rank (MRR), Recall@5 (R@5), and Recall@20 (R@20) are used to evaluate first stage conversational retrieval.

% \paragraph{Wikipedia Corpus}

\subsection{SimSeek for Semi-supervised Setup}
\label{appendix:ssl_detail}

For semi-supervised CQA setup, we set QuAC$_{seen}$ as $\mathcal{D}$ and documents in QuAC$_{unseen}$ as $\mathcal{C}$. The number of turns $T$ is set to 6 in our generations.
%However, we utilize original QuAC train set as $\mathcal{D}$ and external Wikipedia corpus we collected as $\mathcal{C}$ for the full augmentation experiment.
For OR-QuAC experiment, we follow the semi-supervised setup since OR-QuAC shares the same document split with the semi-supervised setup. All CQG models are based on T5-large~\cite{raffel2020exploring} model of 770M parameters, and we use 5 for beam size of beam search and 0.98 for top-$p$ value of nucleus sampling~\cite{holtzman2020curious} with 1.2 temperature. We employ the same backbone for the CAF with corresponding CQA student models, RoBERTa-base (125M), RoBERTa-Large (355M), and Longformer-large (435M)~\cite{liu2019roberta, Beltagy2020Longformer}.

\subsection{Baselines for Synthetic CQA Generation}
\label{appendix:baseline_generaiton}
We introduce strong baselines for synthesizing CQA datasets and compare them with our methods.
For a fair comparison, we train all components 
of approaches on the same labeled dataset $\mathcal{D}$, and generate the synthetic dataset $\hat{\mathcal{D}}$ on the unlabeled corpus $\mathcal{C}$.

% \paragraph{Back-translation}
% Back-translation is one of the most widely used methods for data augmentation in NLP fields~\cite{sennrich-etal-2016-improving, fadaee2017data, xie2020unsupervised}.
% We translate an existing question $q$ from the labeled dataset $\mathcal{D}$ into another target language and then translate it back into the source language to obtain a paraphrased question. We set the target language as German (de) while the source language is English (en). 
% Specifically, we use translation models for both directions (en -> de) and (de -> en), which are pre-trained on WMT-19~\cite{ng-etal-2019-facebook}.

% \paragraph{De-contextualization}
% \citet{elgohary2019can} rewrite conversational questions of QuAC into self-contained questions that could be understood without the conversation. 
% Following \citet{kim2021learn}, we consider the resulting dataset, CANARD, as an additional dataset for training CQA models.
% Note that CANARAD$_{train}$ and QuAC$_{seen}$ share the same passages.

\paragraph{PAQ-CANARD}
For the single-turn QA generation, \citet{lewis2021paq} propose PAQ, the pipeline strategy composed of three phases, answer extraction, question generation, and round-trip filtration.
Even though it is not designed to generate context-dependent questions, we generate de-contextualized conversations like CANARD~\cite{elgohary2019can}. 
Thus, we fine-tune every component of the PAQ on CANARD$_{train}$. Then, we include it as one of the baselines leveraging single-turn QA.

\paragraph{PAQ-QuAC}

We construct a baseline by using a straightforward way to extend the single-turn QG framework, e.g., PAQ~\cite{lewis2021paq}, to a conversational setup.
We replace the question generator in PAQ with CQG$_{answer}$ model that also takes the conversation history as input. Different from our SimSeek-sym, the baseline utilizes the original answer extractor model of \citet{lewis2021paq}, which extracts answer candidates regardless of conversational history, i.e. $p_a(a \mid c)$. From a given answer-containing passage $c$, top-$k$ answer candidates are extracted by the model in advance. Then, we randomly take out an answer from the candidates to feed it to the CQG$_{answer}$ at every turns.

\subsection{Baslines in Downstream tasks}
\label{appendix:baseline_downstream}

\paragraph{CQA Models}
After building synthetic CQA datasets upon the unlabeled corpus $\mathcal{C}$, the baseline CQA models are trained on the datasets.
By comparing the resulting CQA performances, we evaluate the effectiveness of the generated dataset $\hat{\mathcal{D}}$.
We test three backbone architectures for CQA, base
and large size of RoBERTa \cite{liu2019roberta}, and Longformer-large \cite{Beltagy2020Longformer}.
By contrasting various sizes of pre-trained models, we show the different effects of data augmentation.
In addition, we involve Longformer architecture that has been shown to be effective for encoding much longer history \cite{zhao2021ror}, which achieves competitive performance with the state-of-the-art approach.

\paragraph{Conversational Search}

We employ dual-encoder based dense retriever, DPR, for our baseline~\cite{karpukhin2020dense}. Especially, we initialize the encoders with pre-trained DPR model on Natural Questions~\cite{kwiatkowski2019natural}. To represent query input, we concatenate questions $\{q_1, q_2, ..., q_{t}\}$ with \texttt{[SEP]} token. We truncate the input length when longer than 128 but retain first question $q_1$ at the same time~\cite{qu2020open}. The context input is concatenation $c$ and its title with \texttt{[SEP]}. The maximum length for the context input is 384. We train the model for 10 epochs with 128 for batch size, 3e-5 for lr, 0.1 for lr warming up, and 0.01 for weight decay. All DPR models are trained by using in-batch negative without any usage of hard negatives~\cite{karpukhin2020dense}.

\section{Additional Experiments}

\input{tables/unseen_filter}
\input{tables/intrinsic}

\subsection{Intrinsic Evaluation of CQG models}
\label{appendix:instrinsic}

Table \ref{table:quac_gen} presents intrinsic evaluation results of our two kinds of CQG models.
Scores represent the lexical similarity of the generated questions with the ground-truth questions when ground-truth conversational history is given. The sub-component of \textsc{SimSeek-sym}, CQG$_{answer}$ significantly outperforms CQG$_{prior}$ in BLEU scores of all n-gram levels. 
The contrasting results to our experiments (Section \ref{subsec:in_domain_experiment}) imply that accurate generation grounded on the answer is not enough to generate realistic conversation. 
% since the ground-truth history and target answer are given in this setting. 
Instead, we presume other vital factors, such as question based on information asymmetry, proper answer selection in natural conversational flow, and their chained interactions, contribute to a better synthetic CQA generation.

\subsection{Ablation Study on the Filtration}
\label{appendix:ablation_filter}

We perform an ablation study on the filtering process and compare \textsc{SimSeek-sym} to the strong baseline, PAQ-QuAC.
The two frameworks show similar performance in Table~\ref{table:semi_quac}.
Despite the small gap, we observe the end-CQA performances are often saturated by the filtering procedure.
Without the filtration ($\hat{\mathcal{D}}_\text{unfilt}$), \textsc{SimSeek-sym} consistently outperforms PAQ-QuAC over all backbone baselines.
It also shows better filtration efficiency by a gap of approximately 20\%$_\text{p}$ in the success rate.
The results indicate that \textsc{SimSeek-sym} greatly advances the generation frameworks for simulating CQA datasets.

\section{Implementation Details}
\label{appendix:implementation_details}

All our implementations are based on huggingface's transformers library~\cite{wolf2019huggingface}.
% and NAVER Smart Machine Learning (NSML) platform~\cite{sung2017nsml, kim2018nsml}

\subsection{Training models in SimSeek}
\label{appendix:training_simseek}

We train overall four models, CAE, CQG$_{answer}$, CQG$_{prior}$, and CAF on QuAC$_{seen}$ split for SimSeek. We optimize all models using AdamW optimizer with linear learning rate scheduling~\cite{kingma2017adam}. The best-performing checkpoint is selected according to validation score.

We employ 2D span extraction model proposed in PAQ with \texttt{bert-base-uncased} backbone for CAE~\cite{devlin2019bert, lewis2021paq}. We observe that using the previous question and answer pair, $(q_{t-1}, a_{t-1})$, instead of the whole history $\mathcal{H}_t$ is enough to get reasonable performance. The qa pair is appended to $c$ with \texttt{[SEP]} token for input representation. We set overall maximum sequence length to 512 and the maximum history length 32. We train it for 3 epochs with 8 for batch size, 3e-5 for lr on 1 32GB V100 GPU. To evaluate the model, we check whether the ground-truth answer span $a_t$ is in the predicted top-10 answer spans $\hat{A}_t$, i.e. Recall@10.

For both CQG models, we employ same t5-large backbone but different input representations. First, $c$, \texttt{<sep>}, $q_1$, \texttt{<sep>}, $a_1$, ..., $a_{t-1}$, \texttt{<mask>}, $a_t$, \texttt{<sep>} are concatenated to represent input for CQG$_{answer}$, where the \texttt{<sep>} and <mask> are special seperator and masking token, respectively. And the output representation of it is concatenation of \texttt{<bos>}, $q_t$, and \texttt{<eos>}. As mentioned in Section~\ref{sec:simseek-sym}, the $c$ is highlighted by \texttt{<hl>} tokens to emphasize rationale for $a_t$~\cite{gu2021chaincqg}. Second, $\mathcal{B}$, \texttt{<sep>}, $q_1$, \texttt{<sep>}, $a_1$, ..., $a_{t-1}$, <mask> are concatenated to represent input for CQG$_{prior}$ and the output representation is the same with that of $CQG_{answer}$. Actually, the $\mathcal{B}$ is composed of three textual inputs, title, section title, and background (abstractive description) \footnote{~Please see Table~\ref{table:case_study}}. They are also concatenated with the \texttt{<sep>} token to represent the $\mathcal{B}$. The masked question prediction scheme is inspired by \citet{chada2021fewshotqa} and we find the scheme is more sample efficient in our preliminary experiment. We train both CQG models for 10 epochs with 16 for batch size, 3e-5 for lr, 0.1 for lr warming up, and 0.01 for weight decay on 2 32GB V100 GPUs. We set maximum sequence length for the input representations to 512 and maximum context length to 384. The context means $c$ and $\mathcal{B}$ for CQG$_{answer}$ and CQG$_{prior}$, respectively.

We adopt three CQA backbone architectures, RoBERTa-base, RoBERTa-large~\cite{liu2019roberta}, and Longformer-large~\cite{Beltagy2020Longformer}, which are shown to be effective in CQA task. Please note that any CQA models can be used for CAF model as a teacher.
For all CQA models, we concatenate the title, sub-title, and previous history to question text, separating with the special token \texttt{[SEP]}.
We train all models for 2 epochs without weight decay on all datasets and set maximum answer length 64.
CQA models return \texttt{CANNOTANSWER} when all scores of answer logits do not exceed a pre-defined threshold.
RoBERTa backbones are trained for batch size 12 per each GPU without weight decay.
We set the maximum length for query and input sequences as 128 and 512, respectively.
Due to their limitation of the input sequence length, a single question-answer pair at previous turn $(t-1)$ is included to the input, shown to be most effective in prior works~\cite{qu2019bert}.
When Longformer architecture is adopted, we find the optimal setup of the maximum length for query and sequence as 768 and 2048, respectively.
It encodes all previous history and titles when providing answers.
They are trained with batch size 1 per GPU.
For the large-size models, we train them with a learning rate 1.5e-5 on 8 32GB V100 GPUs.

\subsection{Training on Wiki-SimSeek}
When trained on \textsc{Wiki-SimSeek}, we validate models every 20000 steps with the validation split of QuAC.
We evaluate best performing models on the development set

\section{Dataset Statistics}
\label{appendix:data_stat}

\input{tables/data_stat}

Table~\ref{table:data_stat} shows data statistics used in our experiments.

\subsection{Case Study}
\input{tables/qualitative}

We explore how SimSeek-sym fails to simulate realistic conversations, but SimSeek-asym successfully mimics information-seeking behaviors.

The first case in Table \ref{table:case_study} shows the synthetic conversation simulated by SimSeek-sym.
In the example, consecutive and disjoint spans are selected for answers from the evidence document as the conversation progresses.
Moreover, all questions contain a common phrase ``What happened $\cdots$  '' while mentioning keywords that have appeared in previous answers.
Asking these ambiguous questions repeatedly would be the best option for the answer-grounded CQG to achieve answer relevance and coherence with the conversation easily.

On the other hand, we observe various information-seeking behaviors in the second case from our SimSeek-asym.
The lack of information impels questioners to ask open-ended questions using uncertain words such as ``some of the things'' in $q_4$.
When they cannot find adequate follow-up questions on the conversation, they ask an additional information as in $q_5$.
They sometimes fail to acquire new knowledge when the question cannot be answered by the evidence document (see $q_6, a_6$).

\section{Human Evaluation Details}
\label{app:mturk}

\begin{figure*}[t!]
    \centering
    \includegraphics[width=\textwidth]{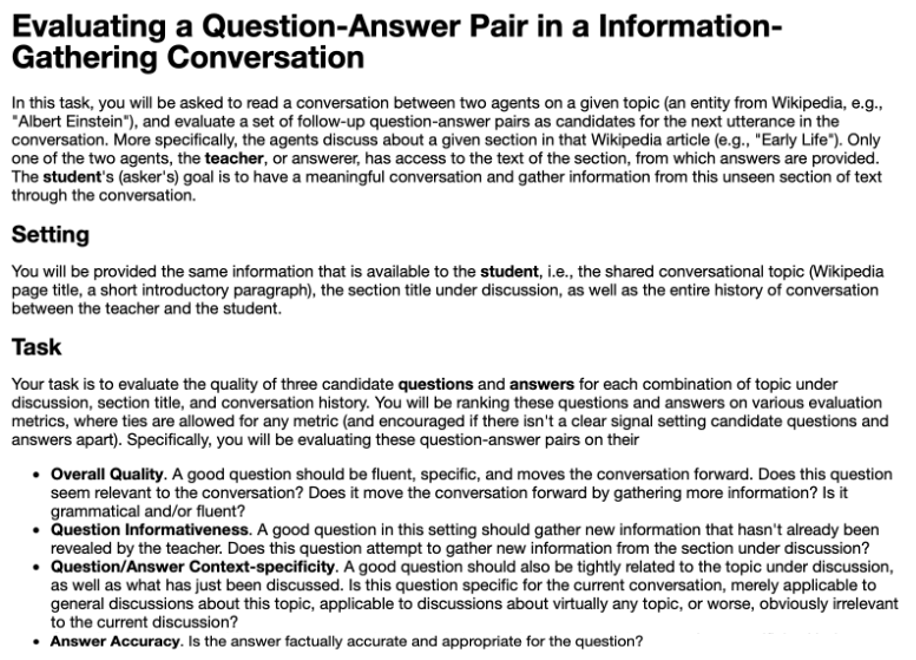}
    \caption{The detailed instructions given to the crowdworkers during human evaluation on Amazon Mechanical Turk. The task description and the settings were based on prior work \cite{qi2020stay}}
    \label{fig:mturk-instruct}
\end{figure*}

\paragraph{Data Preparation}
We ask five workers to compare and rate each candidate follow-up QA generated by one of the models or sampled from the original dataset, given the dialogue history and document context (total 296 samples).
The workers were asked to assess the five criteria ranging from the informativeness of the question and the answer accuracy. 
% The details of the criteria and instructions are described in Appendix \ref{app:mturk}.

As reported by \citet{li2021ditch}, we also observe that the annotators exhibit a bias for questions that do not have an answer
(i.e., \texttt{CANNOTANSWER}).
In addition, we find that the annotators tend to score favorably ``Anything else?'' questions in most criterion  since they often seem relevant to their answer and conversational context.
Thus, we filtered out those types of QA pairs when reporting the scores.

\section{Other Details}
\paragraph{Computational Cost}
We conduct training and inference once for all experiments since it takes huge computational cost. 
Training Longformer-large on \textsc{Wiki-SimSeek} takes 4 days for machines with 8 V100 GPUs.

%% file: tables/unseen_filter.tex
\begin{table*}[t!]
    \small
    \centering
    \begin{threeparttable}
    \begin{tabular}{l|cc|cc|cc|cc}
        \toprule
        Synthetic CQA & \multicolumn{2}{c|}{Filtration} & \multicolumn{2}{c|}{RoBERTa-base} &  \multicolumn{2}{c|}{RoBERTa-large}  & \multicolumn{2}{c}{Longformer-large} \\
        Generation & \#($\hat{\mathcal{D}}$) & \%(Success) & $\hat{\mathcal{D}}_\text{unfilt}$ & $\hat{\mathcal{D}}$ & $\hat{\mathcal{D}}_\text{unfilt}$ & $\hat{\mathcal{D}}$ & $\hat{\mathcal{D}}_\text{unfilt}$ & $\hat{\mathcal{D}}$ \\
        \midrule
        Human ($\hat{\mathcal{D}}$ = QuAC$_\text{unseen}$) & 37,753 & - & - & 65.3 & - & 65.0 & - & 72.3 \\
        \midrule
        % PAQ-CANARD & 28,480 & 67.9 \% & 38.2 & 64.3 & 38.8 & 66.5 & 37.5 & 71.5 \\
        PAQ-QuAC & 11,794 & 28.5 \% & 44.9 & \textbf{55.9} & 47.1 & 51.5 & 42.7 & \textbf{61.7} \\
        \textsc{SimSeek-sym} & 19,550 & 46.6 \% & \textbf{51.8} & 55.5 & \textbf{53.3} & \textbf{54.3} & \textbf{53.6} & 60.8 \\
        % \textsc{SimSeek-asym} & 40,028 & 95.4 \% & \textbf{62.5} & \textbf{65.3} & \textbf{64.8} & 67.6 & \textbf{71.5} & \textbf{73.1} \\
        \bottomrule
    \end{tabular}
    \end{threeparttable}
    \caption{Comparison over two baselines with detailed statistics for the filtering process. $\hat{\mathcal{D}}_\text{unfilt}$ represents the CQA dataset that is not filtered by our filtering process. Although the two frameworks achieve similar performances in Table \ref{table:semi_quac}, \textsc{SimSeek-sym} largely outperforms PAQ-QuAC in the no-filtering setup.}
    \label{table:semi_nofilter}
\end{table*}

%% file: tables/intrinsic.tex
\begin{table}[t!]
    \centering
    \small
    \begin{threeparttable}
    \begin{tabular*}{0.98\columnwidth}{llcccc}
        \toprule
        Trained on & Model & B-1 & B-2 & B-3 & B-4 \\
        \midrule
        \multirow{2}{*}{QuAC$_{seen}$} & CQG$_{answer}$ & 28.2 & 18.0 & 11.9 & 9.1 \\
        & CQG$_{prior}$ & 23.9 & 14.4 & 9.0 & 6.4 \\
        \midrule
        \multirow{2}{*}{QuAC$_{full}$} & CQG$_{answer}$ & 29.6 & 19.3 & 12.8 & 9.7 \\
        & CQG$_{prior}$ & 24.5 & 15.2 & 9.8 & 7.4 \\
        \bottomrule
    \end{tabular*}
    \end{threeparttable}
    \caption{Automatic evaluation over two different CQG models of our frameworks on QuAC development set. The B-* indicate BLEU scores.}
    \label{table:quac_gen}
\end{table}

%% file: tables/data_stat.tex
\begin{table}[t!]
    \centering
    \small
    \begin{threeparttable}
    \begin{tabular*}{0.95\columnwidth}{l|rrrr}
        \toprule
        \textit{QuAC} & \multicolumn{3}{|c|}{Train} & Dev \\
         \midrule
         \textit{QuAC}$_{split}$ & \multicolumn{1}{c|}{Seen} & \multicolumn{1}{c|}{Unseen} & \multicolumn{1}{c|}{Valid} & Dev \\
        \midrule
        \# Passages & 4,383 & 6,694 & 490 & 1,000 \\
        \# Questions & 31,527 & 37,753 & 3,430 & 7,354 \\
        \midrule \midrule
        \textit{OR-QuAC} & \multicolumn{1}{c|}{Seen} & \multicolumn{1}{c|}{Unseen} & \multicolumn{1}{c|}{Dev} & Test \\
        \midrule
        \# Passages & 4,383 & 6,694 & 490 & 771 \\
        \# Questions & 31,527 & - & 3,430 & 5,571 \\
        \bottomrule
    \end{tabular*}
    \end{threeparttable}
    \caption{Data statistics of QuAC dataset used in our experiments. Note that we use questions and answers in QuAC$_{unseen}$ to represent human upper bound. OR-QuAC also contains 11M of chunked passages collection for the retrieval. We split datasets following CANARD~\cite{elgohary2019can}, which is smilar with OR-QuAC \cite{qu2020open}}
    \label{table:data_stat}
    \vspace{-3mm}
\end{table}

%% file: tables/qualitative.tex
\begin{table*}[t!]
\centering
\resizebox{\textwidth}{!}{
\begin{tabular}{l}
\toprule
\large \textbf{SimSeek-sym} \par \\
% C_ac3c9865466046609ebfb8543b6a5ee5_0
\textbf{Title} : Native Americans in the United States \quad \textbf{Section Title} : Self-determination \\ 
\midrule
\textbf{Document $c$} : \\
$\cdots$ \\
Upset with tribal government and the failures of the federal government to enforce treaty rights, \colorbox{yellow!40}{about 300 Oglala Lakota} \\ 
\colorbox{yellow!40}{and AIM activists took control of Wounded Knee on February 27, 1973.} \colorbox{blue!40}{Indian activists from around the country joined}\\
\colorbox{blue!40}{them at Pine Ridge, and the occupation became a symbol of rising American Indian identity and power.} \colorbox{red!40}{Federal law} \\
\colorbox{red!40}{enforcement officials and the national guard cordoned off the town, and the two sides had a standoff  for 71 days.} \\ $\cdots$
\\ \midrule
$\cdots$ \\
\textbf{$q_\text{3}$} : What happened at Wounded Knee? \\
\colorbox{yellow!40}{\textbf{$a_\text{3}$}} : Indian activists from around the country joined them at Pine Ridge, and the occupation became a symbol of rising \\
American Indian identity and power. \\
\textbf{$q_\text{4}$} : What happened after they took control of Pine Ridge? \\
\colorbox{blue!40}{\textbf{$a_\text{4}$}} : Federal law enforcement officials and the national guard cordoned off the town, and the two sides had a standoff \\
for 71 days. \\
\textbf{$q_\text{5}$} : What happened during the standoff? \\
\colorbox{red!40}{\textbf{$a_\text{5}$}} : During much gunfire, one United States Marshal was wounded and paralyzed. \\
$\cdots$ \\
\midrule \midrule
\large \textbf{SimSeek-asym} \par \\
\textbf{Title} : Thor Heyerdahl \quad \textbf{Section Title} : Kon-Tiki expedition\\ \midrule
% C_9020b7dd5db84df1ad401506ed642d37_1
\textbf{Background} $\mathcal{B}$ \\
Thor Heyerdahl (October 6, 1914 - April 18, 2002) was a Norwegian adventurer and ethnographer with a background \\
in zoology, botany, and geography. He became notable for his Kon-Tiki expedition in 1947,
$\cdots$ \\
\midrule
$\cdots$ \\
\textbf{$q_\text{4}$} : What were \underline{some of the things} he found on the Kon-Tiki expedition? \\
\textbf{$a_\text{4}$} : The raft proved to be highly manoeuvrable, and fish congregated between the nine balsa logs in such numbers that \\ ancient sailors could have possibly relied on fish for hydration in the absence of other sources of fresh water.  \\
\textbf{$q_\text{5}$} : Are there any other interesting aspects about this article? \\
\textbf{$a_\text{5}$} : The documentary film of the expedition entitled Kon-Tiki won an Academy Award in 1951.
 \\
\textbf{$q_\text{6}$} : Why did the film win an Academy Award? \\
\textbf{$a_\text{6}$} : CANNOTANSWER \\
\bottomrule
\end{tabular}}
\caption{Examples of the resulting datasets simulated by SimSeek-sym and SimSeek-asym. In the first case (above), SimSeek-sym asks unspecific questions repeatedly, which can effortlessly achieve the goals, answer relevance and coherence with the conversation; but it leads to the shallow conversation.
On the contrary, SimSeek-asym successfully mimics diverse information-seeking behaviors that are commonly occurred in human dialogue.
}
\label{table:case_study}
\vspace{-.4cm}
\end{table*}

\begin{table*}[t!]
\centering
\resizebox{\textwidth}{!}{
\begin{tabular}{l}
\toprule
\textbf{Title} : Esports \quad \textbf{Section Title} : History Early history (1972–1989) \\ 
\midrule
\textbf{Document $c$} \\
\colorbox{red!40}{The earliest known video game competition took place on 19 October 1972 at Stanford Universit}\\ \colorbox{red!40}{for the game "Spacewar".} Stanford students were invited to an "Intergalactic spacewar olympics"\\ whose grand prize was a year\'s subscription for "Rolling Stone", with \colorbox{green!40}{Bruce Baumgart winning} \\\colorbox{green!40}{the five-man-free-for-all tournament and Tovar and Robert E. Maas winning the Team Competition.}\\ \colorbox{blue!40}{The Space Invaders Championship held by Atari in 1980 was the earliest large scale video game competition,}\\ \colorbox{blue!40}{attracting more than 10,000 participants across the United States}, establishing competitive gaming \\as a mainstream hobby.
$\cdots$ \\
\midrule
\textbf{Background} $\mathcal{B}$ \\
Esports (also known as electronic sports, e-sports, or eSports) is a form of competition using video games. \\ Most commonly, esports takes the form of organized, multiplayer video game competitions,\\ particularly between professional players, individually or as teams.\\ Although organized online and offline competitions have long been a part of video game culture,\\ these were largely between amateurs until the late 2000s, when participation by professional gamers and\\ spectatorship in these events through live streaming saw a large surge in popularity.\\ By the 2010s, esports was a significant factor in the video game industry,\\ with many game developers actively designing toward a professional esports subculture. \\
\midrule
\textbf{Conversation in \textsc{Wiki-SimSeek}} \\
\textbf{$q_\text{1}$} : \textbf{What is the history of esports?} \\
\textbf{$a_\text{1}$} : \colorbox{red!40}{The earliest known video game competition took place on 19 October 1972 at Stanford University}\\ \colorbox{red!40}{for the game "Spacewar".} \\
\textbf{$q_\text{2}$} : \textbf{What was the result of this competition?} \\
\textbf{$a_\text{2}$} : \colorbox{green!40}{Bruce Baumgart winning the five-man-free-for-all tournament and Tovar and Robert E. Maas}\\ \colorbox{green!40}{winning the Team Competition.} \\
\textbf{$q_\text{3}$} : \textbf{Did esports grow from there?} \\
\textbf{$a_\text{3}$} : \colorbox{blue!40}{The Space Invaders Championship held by Atari in 1980 was he earliest large scale video game competition,}\\ \colorbox{blue!40}{attracting more than 10,000 participants across the United States,} \\
\textbf{$q_\text{4}$} :\textbf{ What happened after the Space Invaders Championship?} \\
% \textbf{$a_\text{4}$} : In the summer of 1980, Walter Day founded a high score \\record keeping organization called Twin Galaxies.
$\cdots$ \\
\bottomrule
\end{tabular}}
\caption{Another qualitative example in \textsc{Wiki-SimSeek}. Especially, it shows that our framework works well even for topics of out-of-domain, i.e., ``esports'', which is not person-related categories as in original QuAC.
}
\label{table:case_study_ood}
\vspace{-.4cm}
\end{table*}